\documentclass[11pt,a4paper]{article}
\usepackage{geometry}
\usepackage{indentfirst}
\usepackage{graphicx}
\usepackage{subfigure}
\usepackage{algorithmic}
\usepackage{algorithm}
\usepackage{amsmath}

\title{Shared Backbone PPO for Multi-UAV Communication Coverage with Connection Preservation }
\author{Z. Jiang$^{1}$\\
$^{1}$ Zhejiang University}
\date{}

\begin{document}
\maketitle

\noindent
\textbf{Abstract}

This paper proposes a Shared Backbone Proximal Policy Optimization (Shared Backbone PPO) algorithm. By sharing the base module between the Actor and Critic networks, the algorithm achieves efficient training and improved performance. The algorithm is implemented in a connectivity-preserving multi-UAV swarm communication coverage task and compared with the standard PPO algorithm. Experimental results demonstrate that the proposed method achieves superior performance. Furthermore, a graph information aggregation module is incorporated into the model architecture to accommodate the communication conditions among agents. With the integration of this module, the algorithm remains effective, and the trained agent swarm exhibits a higher level of cooperation.



\section{Introduction}
In recent years, UAV swarm networks have attracted extensive attention from both industry and academic research, owing to their potential applications in surveillance, mobile communication systems, wildfire monitoring, traffic monitoring, and so on [4,21]. UAV swarm networks are composed of a large number of UAVs that cooperate with each other to provide communication services to ground or low-altitude members [19]. Benefiting from the sky advantage of UAVs, UAV swarm networks can seek optimal areas in the air to achieve rapid deployment and flexibly avoid various communication obstacles [17], thereby providing optimal signal coverage for terminals while dynamically adjusting the distribution of UAV base stations to maximize service coverage. In certain special missions, the communication links established by UAVs can also guarantee communication security and reliability. Functioning as aerial base stations, UAVs possess a larger coverage range and stronger environmental adaptability, making them less susceptible to interference from ground environmental obstacles compared to terrestrial base stations. For mobile terminal access points, aerial base stations can better accommodate their mobility; by following or orbiting, they can provide stable communication signals for the target terminals.


UAV swarm network coverage tasks [5,21] can be broadly categorized into two types based on the network organization approach: connectivity preservation coverage [3] and standalone coverage [6]. In connectivity preservation coverage, the swarm needs to maintain the distance between UAVs to keep them interconnected, forming a connected graph. The advantage of this networking approach is that it ensures information collected by one UAV node can be transmitted to other nodes, ultimately enabling all nodes connected to the network to interact with each other. In standalone coverage, UAVs do not maintain communication links; instead, they perform coverage independently. This approach is widely applied in wireless sensor networks. It primarily involves UAVs collecting data from low-altitude or ground terminals, temporarily storing it in the aerial base stations, and subsequently retrieving the data centrally to a data center for processing. Although this approach has slightly lower timeliness, the cooperation process of the UAV swarm is easier to implement, and the mobility of each UAV is less constrained. Connectivity preservation coverage imposes higher requirements on inter-UAV cooperation, but it also possesses broader application scenarios. It enables more effective data exchange for real-time data and facilitates remote interaction through aerial relaying [18].

\begin{figure}[htbp]
    \centering
    \includegraphics[width=16cm]{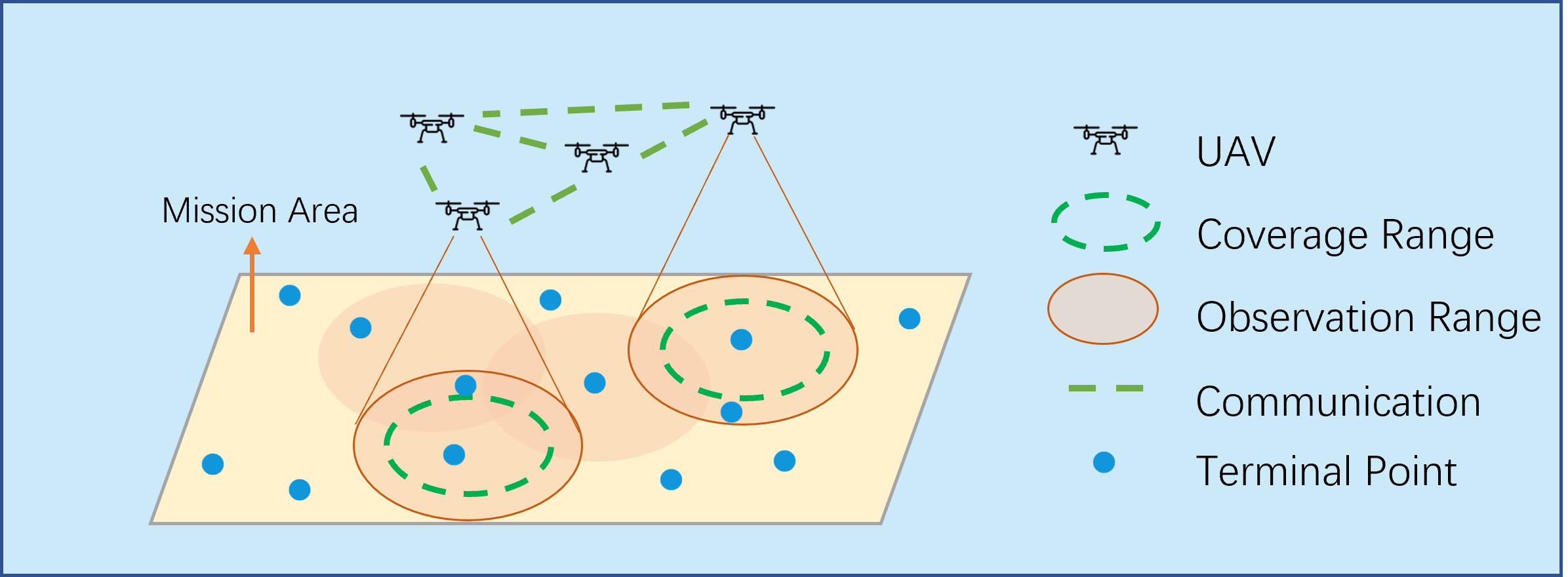}
    \caption{UAV Swarm Networks for Communication Coverage}
    \label{scenario}
\end{figure}

In this paper, we primarily focus on connectivity preservation coverage in UAV swarm network coverage tasks, as shown in Figure \ref{scenario}. The UAV swarm needs to balance the coverage task and the connectivity preservation task. While maintaining connectivity, the swarm attempts to cover as many ground terminals as possible. Under this organizational scheme, a UAV should neither fly out of the communication range of its neighboring UAVs, nor should it stay too close; rather, it should maintain a certain degree of separation to maximize sensing coverage toward the ground terminals. We achieve this challenging cooperative communication scheme through multi-agent reinforcement learning (MARL). Through training, the agents are enabled to achieve autonomous collaboration, fulfilling the communication objective that accommodates both connectivity preservation and maximum coverage.


The main contributions of this paper are as follows:

1)We propose a novel PPO architecture that enables multi-agent swarm systems to better learn coverage policies. By constructing a shared backbone network for the Actor and Critic, it allows the multi-agent swarm to acquire more effective coverage strategies.

2)We design a multi-agent joint coverage scenario to evaluate the impact of the algorithm on enhancing the performance of multi-agent coverage tasks. In this scenario, each UAV is treated as an agent. The UAVs are required to cover as many terminal targets as possible, while the UAV nodes within the swarm must maintain the properties of a connected graph.

3)We conducted a series of experiments to compare the performance of PPO under different structures. Furthermore, we incorporate a graph-based policy into the multi-agent PPO, enabling agents to aggregate neighborhood information and thereby further improving their decision-making performance.





\section{Related Work}
Reinforcement learning (RL) is widely applied in fields such as robotic autonomous navigation [15], autonomous vehicle driving [14], UAV flight decision-making [11], logistics route optimization [13], conversational dialogue systems [12], and advertising recommendation systems [16], among others. Depending on the learning objective, reinforcement learning can be categorized into two types: Value-based reinforcement learning [2] and Policy-based reinforcement learning [1]. These two types of RL algorithms differ significantly in their structural design.

Value-based RL, represented by the Q-learning [9] and DQN [10] algorithms, learns a Q-value function through continuous training, and then utilizes this Q-value function to enable the agent to take actions based on the current situation. The Q-value function can infer different Q-values via a neural network depending on the agent’s state. Each Q-value output consists of a set of Q-values corresponding to every possible action the agent can take, and the agent ultimately decides which action to execute. For the Q-function, this constitutes a deterministic policy; since the weight parameters of the neural network remain unchanged, the values inferred by the network for the same agent observation will not change. However, during the action selection process, the agent employs a certain stochastic strategy, introducing randomness into each interaction with the environment, thereby collecting rich experiential data. In many cases, the agent adopts the 
$\epsilon-$greedy strategy to select actions. Because Value-based RL algorithms need to map values to each action, they require the agent’s actions to be discrete.

Policy-based RL, represented by the SAC [8] and PPO [7] algorithms, aims to train and obtain a policy function. However, the training process is not composed solely of the policy network; these algorithms typically consist of two networks: the Actor and the Critic. The Actor network is responsible for making decisions and generating the probability distribution of actions, while the Critic network is responsible for evaluating the performance level of the agent’s executed actions. The final policy is determined by the Actor network, which outputs the action probability distribution, and the agent samples from this distribution to obtain the action to execute. Precisely because the Actor policy network can output action probability distributions, Policy-based RL algorithms can accommodate continuous action spaces, but the complexity of their training process is higher than that of Value-based RL algorithms. Policy-based RL algorithms possess broad adaptability and are highly favored by researchers due to their ability to control robots with continuous action spaces and handle high-dimensional state spaces and complex tasks.


This paper will adopt the well-established Policy-based PPO algorithm [7] as the control algorithm for the UAV swarm. By utilizing the improved PPO algorithm to control the UAV swarm, the agent swarm can achieve self-organized collaboration to execute tasks. The full name of PPO is Proximal Policy Optimization. It is an on-policy learning algorithm [20], meaning the learning policy and the decision-making policy utilize the same set of neural networks. By employing a clipping mechanism on the loss, the algorithm ensures that the Actor network of the updated new policy does not deviate excessively from the Actor network of the old policy, thereby avoiding instability and policy collapse during training.

The PPO algorithm accomplishes its training objectives through the collaboration of the Actor and Critic networks. First, the Actor network determines the probability distribution of agent actions based on the environmental information observed by the agent. The agent then selects an action to interact with the environment according to this probability. Through multiple interactions, a trajectory is formed. The Critic network evaluates the quality of the actions decided by the Actor based on the states within the trajectory. Subsequently, under the guidance of the Critic, the Actor network continuously optimizes its policy network via policy gradient descent to achieve favorable decision-making performance.

During this process, the agent inputs the observed state information into the Actor network. Upon receiving the state information, the Actor generates a probability distribution over actions. In a discrete action space, the Actor network outputs the probability of each action via a softmax function in the final layer, ensuring that the sum of all action probabilities equals 1. The agent then performs a sampling selection based on this probability. In a continuous action space, the Actor network outputs parameters 
$\mu$ and 
$\sigma$ in the final layer. These parameters form a probability distribution curve, from which the agent samples to select the execution action.

In the PPO algorithm, network optimization relies on trajectories. The agent needs to interact with the environment multiple times, and the network is optimized only after collecting sufficient trajectory data. When optimizing the network, the data is typically used to calculate the Critic network’s metrics first: a trajectory’s state information is input to the Critic network, which then provides an estimated return value based on these states and its own predictions. The Critic’s Loss is calculated using the Mean Squared Error (MSE) between the predicted return and the actual return, 
$Loss=(return\_predict-return\_reality)^2$, which is then used to optimize the Critic network.

The optimization of the Actor network relies on the Critic network, which assists in calculating the $advantage$ function  and combines it with the action probability distribution generated by the Actor to compute the  $Importance Sampling$. The final network optimization loss is obtained by combining both with the clipping strategy. There are multiple methods to generate the  $advantage$ function; the simplest is the Monte Carlo method, $advantage=-return\_predict+return\_reality$, as well as the GAE method.

A single optimization of the Actor network undergoes multiple rounds of network parameter weight updates, and the $Importance Sampling$ needs to be recalculated in each round. The $Importance Sampling$ 
is calculated using the Actor network at the start of the current optimization as the baseline, combined with the current Actor network. First, the old Actor is used to calculate the probability of sampling this action under the state in the original Actor policy; then, the updated new Actor is used to calculate the probability of sampling this action under the state in the current Actor policy, $Importance Sampling=\frac{\pi_{\theta_{\text {new }}}\left(a_{t} \mid s_{t}\right)}{\pi_{\theta_{\text {old }}}\left(a_{t} \mid s_{t}\right)}$. Subsequently, the Importance Sampling is clipped within a specific range. When the sampling weight is too large, the update magnitude is restricted, which limits training instability and policy collapse caused by excessively large updates while maintaining exploration capability,
$Loss\_CLIP = clip(IS,1-epsilon,1+epsilon)*A$,$Loss\_CPI = IS*A$,$Loss = min(Loss\_CPI,Loss\_CLIP)$, $loss = -Loss.mean()$. The final loss used to update the Actor network’s weight parameters is obtained by averaging the multiple losses derived from the multiple actions reflected in the trajectory. The negative sign is applied because the Actor network parameters are updated via gradient descent.

After completing the updates of the Critic and Actor, the experience data in the replay buffer is discarded. The newly generated New Actor then interacts with the environment again to form sufficient trajectory data for the next round of network optimization.

\section{Method(Shared Backbone PPO)}
In this section, we will introduce the model design of the multi-UAV communication coverage task in detail, and then present our multi-agent reinforcement learning policy model, Shared Backbone PPO. By adopting a shared backbone approach for the Actor and Critic networks within the PPO model, our algorithm achieves better performance compared to the independent Actor-Critic architecture PPO. We apply both the traditional independent Actor-Critic architecture PPO and the shared-backbone Actor-Critic architecture PPO to the multi-UAV communication coverage task, comparing the adaptability and performance differences of the algorithms under different architectures. Meanwhile, we also employ the agent neighborhood information aggregation technique, which further enhances the performance of the algorithm.


\subsection{System Model}
\subsubsection{Coverage Scenario}
In this section, we propose the UAV communication coverage task and detail the scenario design of our multi-UAV cooperative communication coverage. Our UAV swarm needs to cover as many ground terminal nodes as possible within a limited number of time steps. Simultaneously, the UAVs must satisfy the connectivity preservation constraint to ensure that the terminal nodes within the communication network can successfully exchange information. The simulation world is a 2D map where we randomly generate 120 communication-demand terminals. Under the condition of maintaining a connected topology, our UAVs must cooperate to cover these demand terminals as extensively as possible to provide necessary services for users. Each UAV has a specific communication coverage range; only the terminal nodes within this range can receive communication services from the UAV. Meanwhile, each UAV has a specific observation range, allowing it to observe the demand terminals within this area and navigate towards them to provide services. Furthermore, there is a specific communication capability between UAVs. UAVs must operate within this inter-UAV communication range to ensure they do not act in isolation and disrupt the connectivity of the swarm. At the same time, UAVs can utilize this intra-swarm communication network to exchange information, communicating with neighbors about their own observations and acquiring information observed by neighbors, thereby achieving better swarm cooperation.


\subsubsection{Observation Space and Action Space}
In the UAV swarm, each UAV serves as an agent, and each agent has its own Observation Space and Action Space. The data within the Observation Space is fed into the policy network to generate decisions, and the resulting decisions fall within the Action Space. The agent’s Observation Space consists of several components, including the UAV’s identification number, its own position information, velocity information, and the content perceived within the UAV’s field of view. When generating the agent’s state information, we apply binary encoding to the UAV’s position information, providing a 20-unit encoding length to represent the agent’s position along both the x-axis and y-axis. The UAV’s velocity information is similarly represented by its velocity components along the x-axis and y-axis. The UAV’s observation information is composed of the elements within the agent’s perception domain: the agent can perceive the position information of ground terminal users within its observation range, as well as the position information of other UAVs within its nearby perception range. These elements are concatenated together to form the Observation Space for the agent’s decision-making task. At each time step, after obtaining the observation information, each agent outputs an adjacency matrix to represent the communication adjacency relationship among the agents in the swarm. The adjacency matrix is an n×n matrix that records the one-hop neighbor relationships among the n agents. If a communication link can be established between UAVs within the communication range, the corresponding entry in the communication relationship matrix is set to 1; otherwise, the matrix element is 0. Each UAV has a 17-dimensional Action Space, which controls the UAV’s acceleration in 8 directions: East, West, South, North, Southeast, Northeast, Southwest, and Northwest. Each direction has two magnitudes of acceleration, plus an additional action of no acceleration in any direction, making a total of 17 action behaviors.


\subsubsection{Evaluation Metrics}

The task performance of the UAV swarm communication coverage is evaluated by considering both the coverage of ground terminal users by the UAVs and the energy consumption of the UAVs in the swarm. The coverage condition of ground terminal users by the UAVs consists of two components: individual coverage and collective coverage. The individual coverage reward is determined by the number of terminal nodes covered by the individual UAV. When the UAV does not cover any terminal user, the individual reward is -1, 
$individual\_reward=-1$; in other cases, if the UAV covers 
$n$ terminal nodes, 
$individual\_reward=n$. The collective reward is determined by the average number of terminals covered by the other UAVs excluding the current UAV, 
$group\_reward=0.1*(total\_covered - individual\_covered)$. The sum of the individual reward and the collective reward serves as the final result to evaluate the current agent’s performance state, 
$reward=individual\_reward+group\_reward$. Meanwhile, the power consumption of the UAV is also within the scope of evaluating its performance. The agent’s reward is inversely proportional to the UAV’s power consumption speed; the faster the power consumption, the worse the performance of the agent’s decision-making policy, 
$reward=reward/energy\_consumption$.


\subsection{Shared Backbone PPO Network}
We are dealing with a multi-UAV cooperative control problem here, where the behavior of each UAV affects the decision-making of other UAVs, leading to significant uncertainty in the environment. Such massive randomness can result in unstable policy training. However, in our task, the objective of each UAV is identical: to cooperate with each other to cover as many ground terminal nodes as possible. Furthermore, the observation spaces and action spaces across agents are also identical. Therefore, we adopt a parameter-shared reinforcement learning training scheme, where each agent shares the same set of decision network parameters.


\subsubsection{Graph Aggregator}
We are facing a swarm task. To promote cooperation among agents, we employ a Graph Aggregator that enables agents to fuse their own information with that of their neighboring agents, thereby achieving better swarm cooperation performance. The Graph Aggregator utilizes the adjacency matrix of the swarm to aggregate the information of the agents within it. During this process, the Graph Aggregator first encodes the observation data of the agents to reduce the data dimensionality and extract feature content from the observation information. It then utilizes the swarm’s adjacency matrix to obtain the topological relationship graph among different nodes. With the aid of this relationship graph, it selectively aggregates information from surrounding neighbors. The relationship graph matrix consists of two types of elements: 1 represents a direct neighbor relationship between two agents, while 0 indicates that two agents cannot achieve one-hop communication and have no direct neighbor relationship. Note that an agent is always considered connected to itself. We use $\Vec{x}$
to denote the feature vector of the agent’s encoded observation information, and $\Vec{y}$ to denote the feature vector of the agent after aggregating information with neighbor nodes. $\Vec{y}$ can be described as
$\vec{y}=(D^{-1 / 2} A D^{-1 / 2} \vec{x} )$, where 
$\mathbf{A}$ is the adjacency matrix among the agent nodes in the swarm, and $D$ is a diagonal matrix where each element on the diagonal represents the number of one-hop connected neighbors radiating outward from each agent. Ultimately, the agent’s feature vector $\vec{h}$ obtained by the Graph Aggregator is generated by weighting its own feature vector $\vec{x}$ and the aggregated feature vector
$\vec{y}$, expressed as $\vec{h}=\alpha\cdot\vec{x}+(1-\alpha)\cdot\vec{y}$, where 
$\alpha$ is a weighting parameter. The aggregated feature vector $\vec{h}$is then further utilized for decision-making.


\subsubsection{Model Framework}
The network architecture of PPO consists of an Actor network and a Critic network. These two networks operate independently during inference while collaboratively optimizing each other during parameter updates. During the training process, the Actor is responsible for generating the actions executed by the agents, whereas the Critic evaluates the quality of the interactions between the generated actions and the environment.

In single-agent PPO, the Actor-Critic architecture has only one configuration. However, in multi-agent cooperative tasks, different Actor-Critic configurations can be constructed according to different input states. In the basic setting, PPO processes each agent in the swarm independently. The Actor receives the observation information of the agent, while the Critic directly receives the observation information of the same agent. The Actor generates a series of action trajectories for the agent, and the Critic evaluates each time step within these trajectories, after which the two networks are iteratively optimized. Under this architecture, the algorithm can only focus on the information of an individual agent without considering the concept of the swarm.

\begin{figure}[htbp]
    \begin{minipage}[t]{0.3\textwidth}
        \centering
        \includegraphics[width=4.5cm]{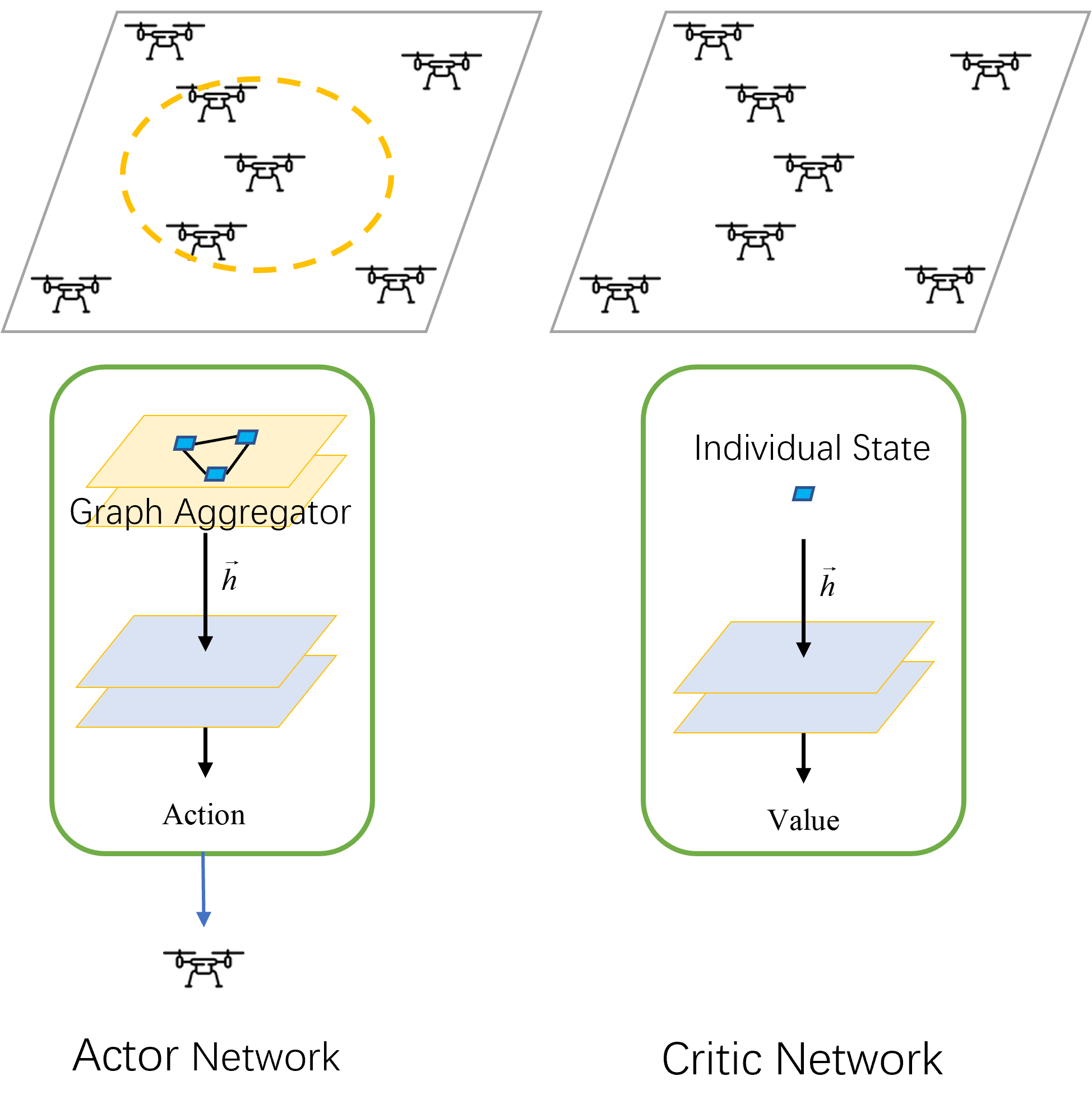}
        \caption{PPO structual 2-1}
        \label{structual2-1}
    \end{minipage}
    \begin{minipage}[t]{0.3\textwidth}
        \centering
        \includegraphics[width=4.5cm]{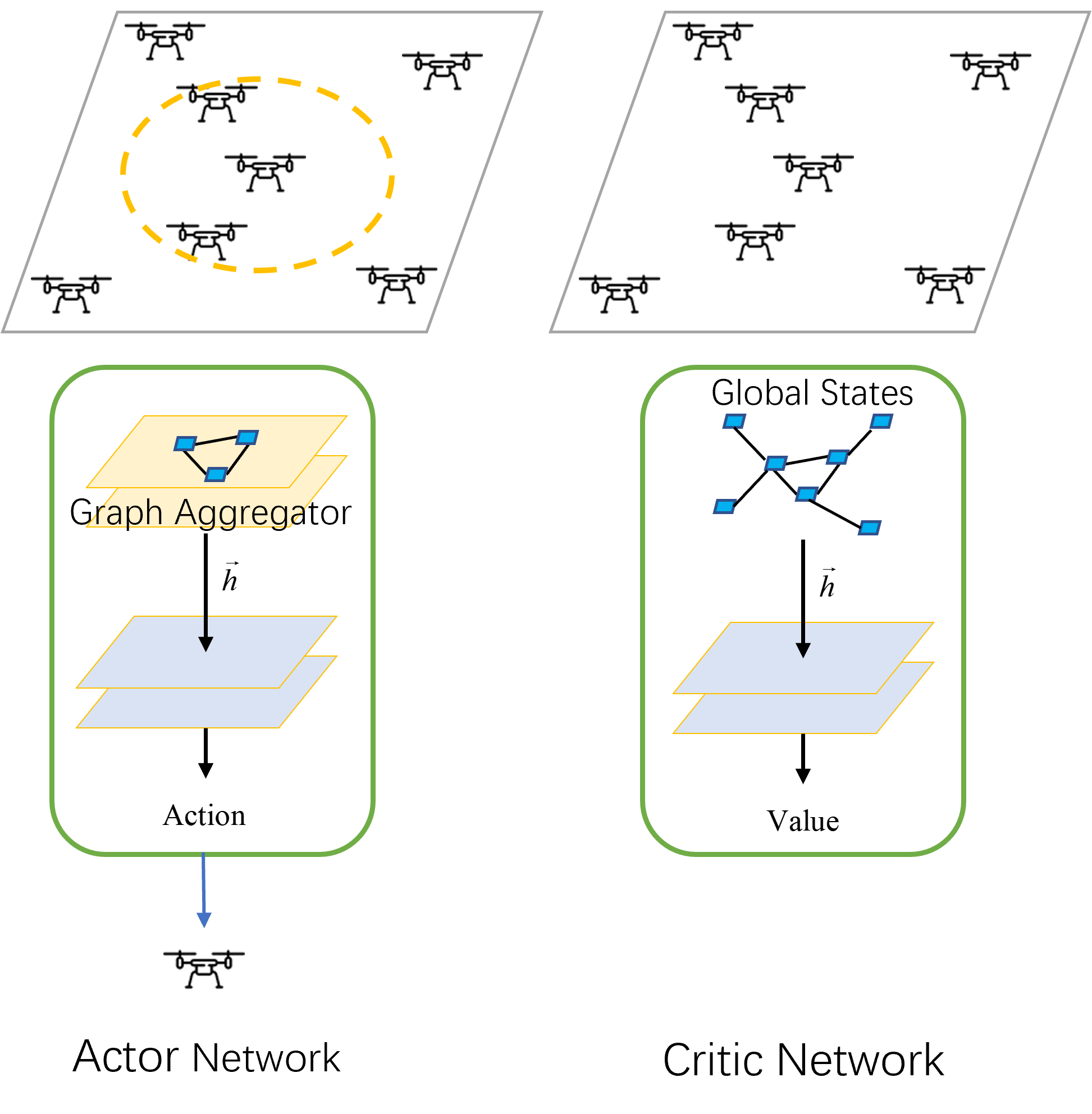}
        \caption{PPO structual 2-2}
        \label{structual2-2}
    \end{minipage}
    \begin{minipage}[t]{0.3\textwidth}
        \centering
        \includegraphics[width=4.5cm]{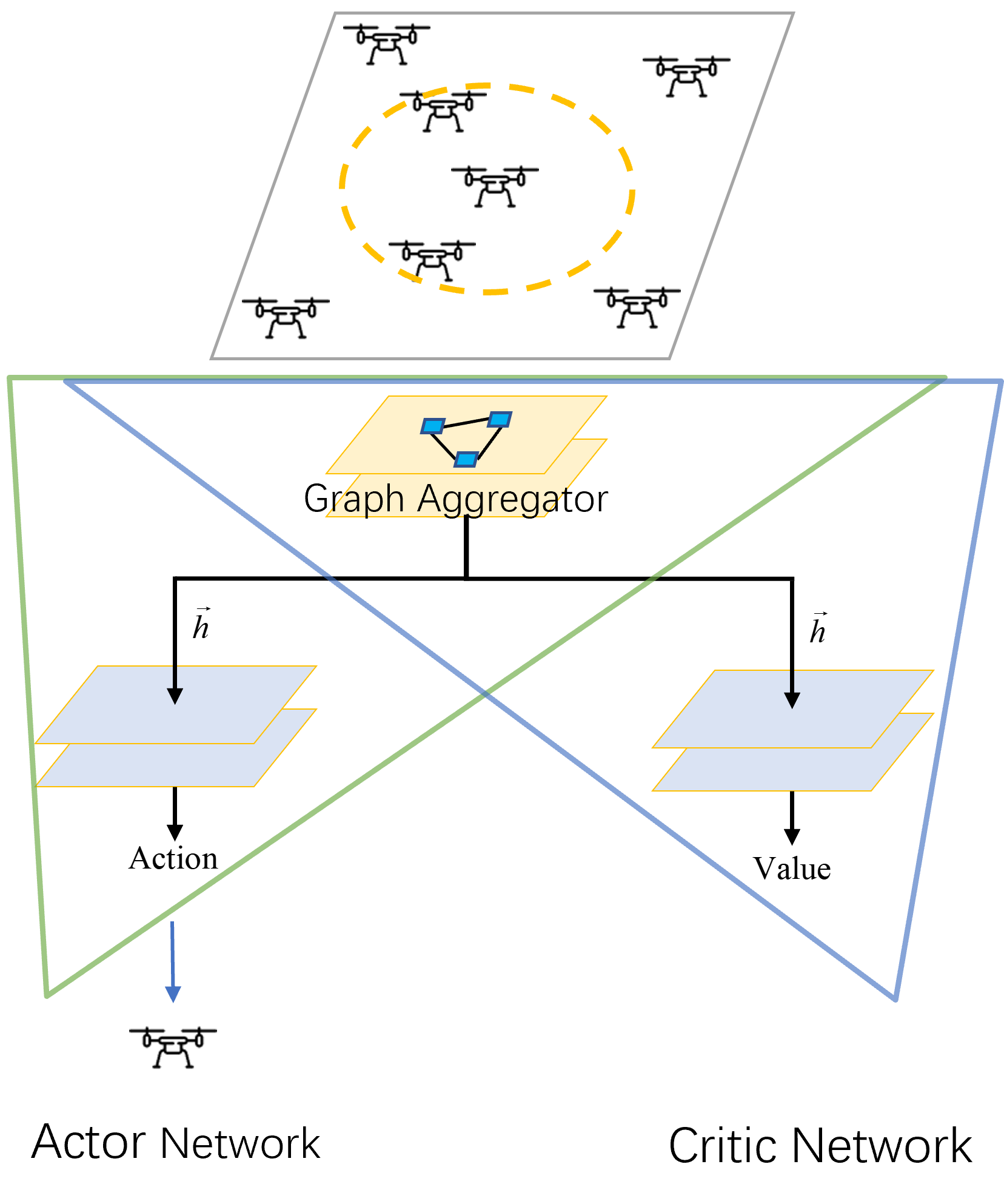}
        \caption{PPO structual 2-3}
        \label{structual2-3}
    \end{minipage}
\end{figure}

In multi-agent swarm scenarios, the architecture can be further adjusted by incorporating a Graph Aggregator, enabling PPO to aggregate information from both the agent itself and its neighboring agents. The aggregated feature vectors are then fed into a Multi-Layer Perceptron (MLP) to infer actions, as shown in Fig.~\ref{structual2-1}. In this way, the Actor can better take neighboring conditions into account during decision-making, thereby adapting more effectively to swarm scenarios.

Furthermore, adjustments can also be made to the Critic in the PPO algorithm, as illustrated in Fig.~\ref{structual2-2}. By leveraging the characteristics of the swarm, the Critic network can receive global information from all agents in the swarm, while the Actor network continues to receive only the local state information surrounding the current agent. This approach enables the Critic to evaluate the actions of agents more reasonably from the perspective of the overall swarm state. However, although the Critic network obtains global swarm information and performs value estimation, its evaluation focus becomes relatively broad and lacks concentrated assessment of the current agent's state value.

To address this issue, we propose the Shared Backbone PPO Network, which is a PPO architecture with a shared backbone between the Actor and Critic. The Actor and Critic networks share the fundamental parameters of the Graph Aggregator module, allowing the Actor to consider neighboring state information during decision-making, while enabling the Critic to focus its attention on the current agent during value estimation. In addition, the Actor and Critic share a common set of parameters, ensuring that behavior decision-making and value evaluation are conducted from a consistent perspective.

During network updates, the backpropagated gradients from the Actor branch loss may conflict with those from the Critic branch loss, since the parameters within the Graph Aggregator are jointly shared by both branches. To resolve this conflict, the shared parameters of the Graph Aggregator are optimized using the gradients propagated from the Critic branch, because the loss of the Critic network is more direct and easier to compute than that of the Actor network.


\subsubsection{Training Process}
Before the beginning of each training episode, the environment is first initialized, after which the agents obtain local observation information, the global observation information of the environment, and the adjacency matrix of the swarm. The agents' observations and adjacency matrix are then fed into the policy network, which consists of an Actor branch and a Critic branch.

The observation information of the agents is first encoded by the Graph Aggregator to obtain feature vectors, which are subsequently passed into the Actor branch to generate the actions of the agents, along with the corresponding action probabilities in the probability distribution. Meanwhile, the Critic branch evaluates the value of the current state according to the current observations of the agents. The selected actions are then executed in the environment for interaction, resulting in the next-step observations, reward values, and updated swarm adjacency matrix. The observation information, reward values, adjacency matrices, as well as the actions, action probabilities, and value estimates corresponding to the current state are all stored in the replay buffer.

After the completion of an episode, the value estimate of the final state is calculated. Based on this final state value estimate and the accumulated rewards, the true return is computed through backward summation. The advantage used to optimize the Actor is then obtained by subtracting the value estimate from the true return. Subsequently, the current Actor network is used to calculate the probability of each action within the trajectory. The ratio between the newly calculated action probability and the action probability stored in the replay buffer forms the importance sampling weight. The Actor loss is then obtained by combining the importance sampling weight, the advantage function, and the clipping operation in PPO.

Meanwhile, the current Critic network estimates the value of each state within the trajectory. The loss of the Critic network is computed based on the difference between the estimated state values and the true returns. This loss is then used to update the parameters of both the Critic network and the Aggregator network, while the loss obtained from the Actor branch is used to update the parameters of the Actor branch.

After the parameter updates are completed, new experience trajectories are collected, and the algorithm enters the next update cycle. This iterative process continues until the learning curve converges.Algorithm 1 describes the process of the Shared Backbone PPO :


\begin{algorithm}[!h]
    \caption{Shared Backbone PPO}
    \label{alg:AOS}
    \begin{algorithmic}[1]
        \STATE  Initialize weights for aggregator, actor and critic networks
        \FOR{episode $epi$ = 1 to T}
            \STATE For each agent $i$, set an empty trajectory $\tau=[]$
            \STATE Reset the environment and initialize the observation $obs$ for each agent $i$
            \FOR{time steps $t$ = 1 to episode length}
            
                \FOR{agent $i = 1$ to N}
                    \STATE $h_{i} \leftarrow Aggregator(obs,adj)$
                    \STATE Select $action_i$ in state $h_{i}$ via  $Actor(action_i \mid h_i)$
                    \STATE Calculate $value$ in state $h_{i}$ via  $Critic(h_i)$
                    \STATE Execute agent’s action $action_i$ to the environment
                \ENDFOR
                \STATE Collect the action and probability, reward, observation, relationships between agents, value 
                \STATE Push the experience into trajectory $\tau += [obs,adj,action,action\_prob,reward,value]$
                \STATE $obs=obs', adj=adj'$
            \ENDFOR
            \STATE Computes $advantage$ and $return$ using trajectory $\tau$
            \FOR{update epoch $epoch$ = 1 to update times}
                \STATE Update weights for actor network
                \STATE Update weights for critic network 
                \STATE Update weights for aggregator network
            \ENDFOR
        \ENDFOR

    \end{algorithmic}
\end{algorithm}

\section{Experiment}
In this section, we introduce the detailed design of the experiments, including the scenario design for the multi-UAV swarm communication coverage task under Connectivity Preservation constraints. The proposed Shared Backbone PPO model is then applied to the multi-UAV communication coverage task. Through comparative experiments and analysis, the performance improvements achieved by the proposed enhanced model over the original model are evaluated and discussed.

\subsection{Experimental Settings}
Our model is deployed on a platform equipped with Ubuntu 22.04.1 Server, one NVIDIA GeForce RTX 4090 GPU, and one AMD Ryzen 9 5950X CPU. The experiments are conducted under an environment configured with Python 3.7.10, PyTorch 1.12.1, and CUDA 12.2.


In the experiments, we investigate different variants of the PPO Actor-Critic architecture to compare their decision-making capabilities in UAV swarms. Under each architecture, the PPO agent serves as the decision-making entity for the UAV swarm, enabling cooperative communication coverage among multiple UAVs. During the task, the UAVs are required to maintain network connectivity while maximizing coverage of ground terminal nodes to provide communication services.

In the proposed Shared Backbone PPO, the Actor and Critic branches share the parameters of the Graph Aggregator module. The Aggregator encodes and aggregates both the UAVs’ local observations and neighborhood information, which is used by the Actor for decision-making. Simultaneously, the aggregated representation is also passed to the Critic network for state value estimation. By enabling parameter sharing between the Actor and Critic within the swarm-aware aggregation module, the proposed method improves the overall performance of the algorithm in multi-agent swarm scenarios.


\subsection{Simulation Scenarios}
In this section, we provide a detailed description of the multi-UAV communication coverage scenario. We design a simulation environment consisting of 10 UAVs acting as aerial mobile base stations (ABSs). Each UAV is equipped with a communication coverage radius of 15 units, a ground sensing range of 19 units, and an inter-UAV communication range of 30 units. The task is conducted on a rectangular map of size 200 × 200 units.
At the beginning of each episode, the 10 UAVs are randomly deployed at different positions on the map. However, the UAV swarm is initialized under the constraint that the communication range of 30 units ensures a connected graph among all UAVs. Meanwhile, 120 terminal nodes are randomly generated across the map. The objective of the 10 UAVs is to cooperatively maximize the number of covered terminal nodes through coordinated behavior.
Each episode consists of 100 time steps. After 100 steps, the environment is reset, and both the terminal nodes and the UAV swarm are reinitialized for the next episode.


\subsection{Experience Result and Discussion}
In this section, we experimentally validate the proposed Shared Backbone PPO algorithm to demonstrate its effectiveness in enabling cooperative UAV control for communication coverage tasks. We further compare PPO under multiple Actor-Critic architectures to investigate their impact on swarm coverage performance. In addition, we evaluate the effect of the Graph Aggregator module through ablation studies, where we compare the performance differences between models with and without the graph aggregation mechanism.
The performance of the algorithms is evaluated from several aspects, including the convergence of the reward curve, the coverage rate of terminal nodes during task execution (coverage index), and the energy consumption of the UAV swarm during operation (energy index).
In the experimental setup, training is conducted in parallel across five scenarios. The results are averaged every 10 episodes to obtain statistically stable performance trends. Based on these metrics, we conduct a comprehensive comparative analysis of all evaluated models.


\begin{figure}[htbp]
    \begin{minipage}[t]{0.5\textwidth}
        \centering
        \includegraphics[width=7cm]{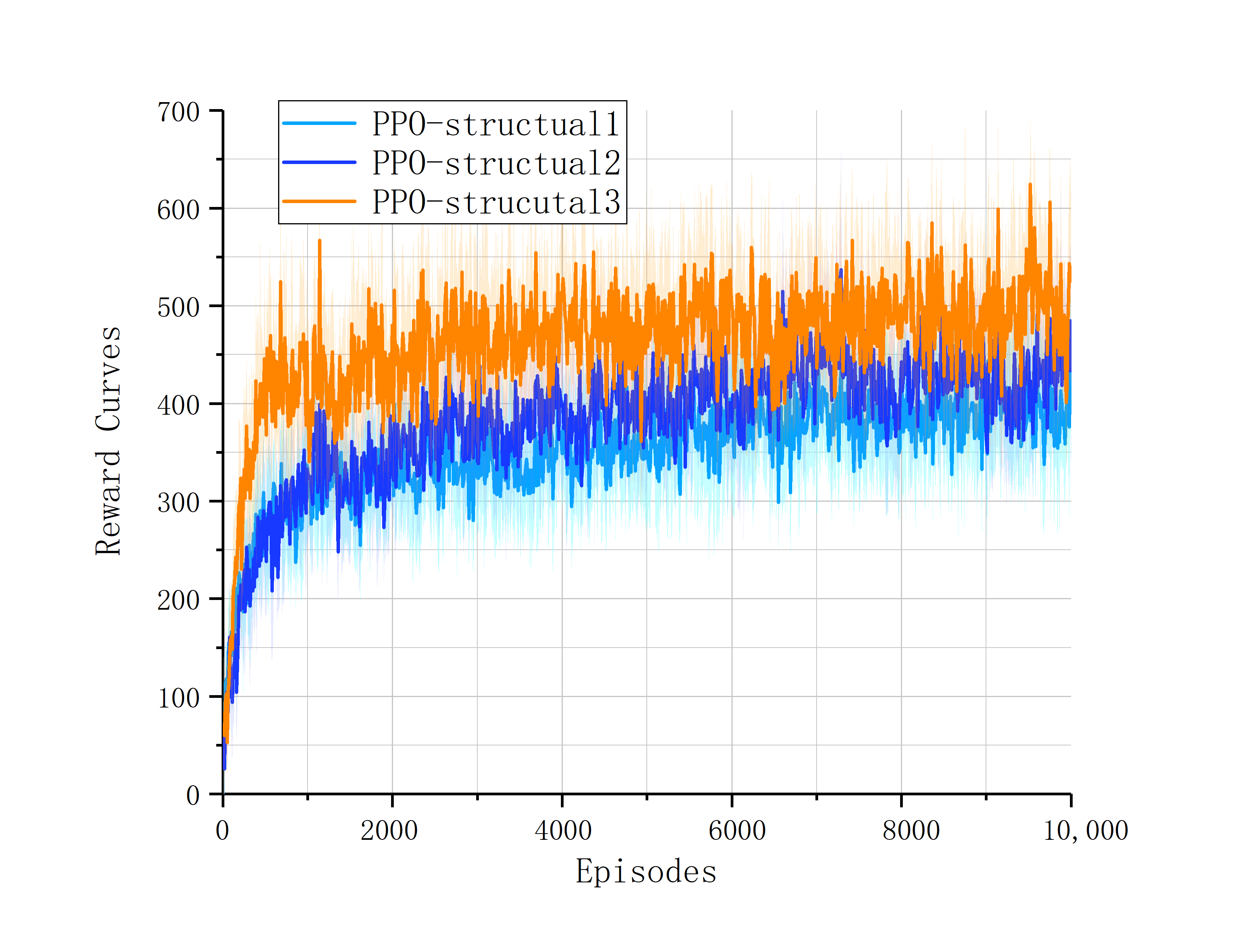}
        \caption{Reward Curves without Graph Aggregator}  
        \label{reward123_curves}
    \end{minipage}
    \begin{minipage}[t]{0.5\textwidth}
        \centering
        \includegraphics[width=7cm]{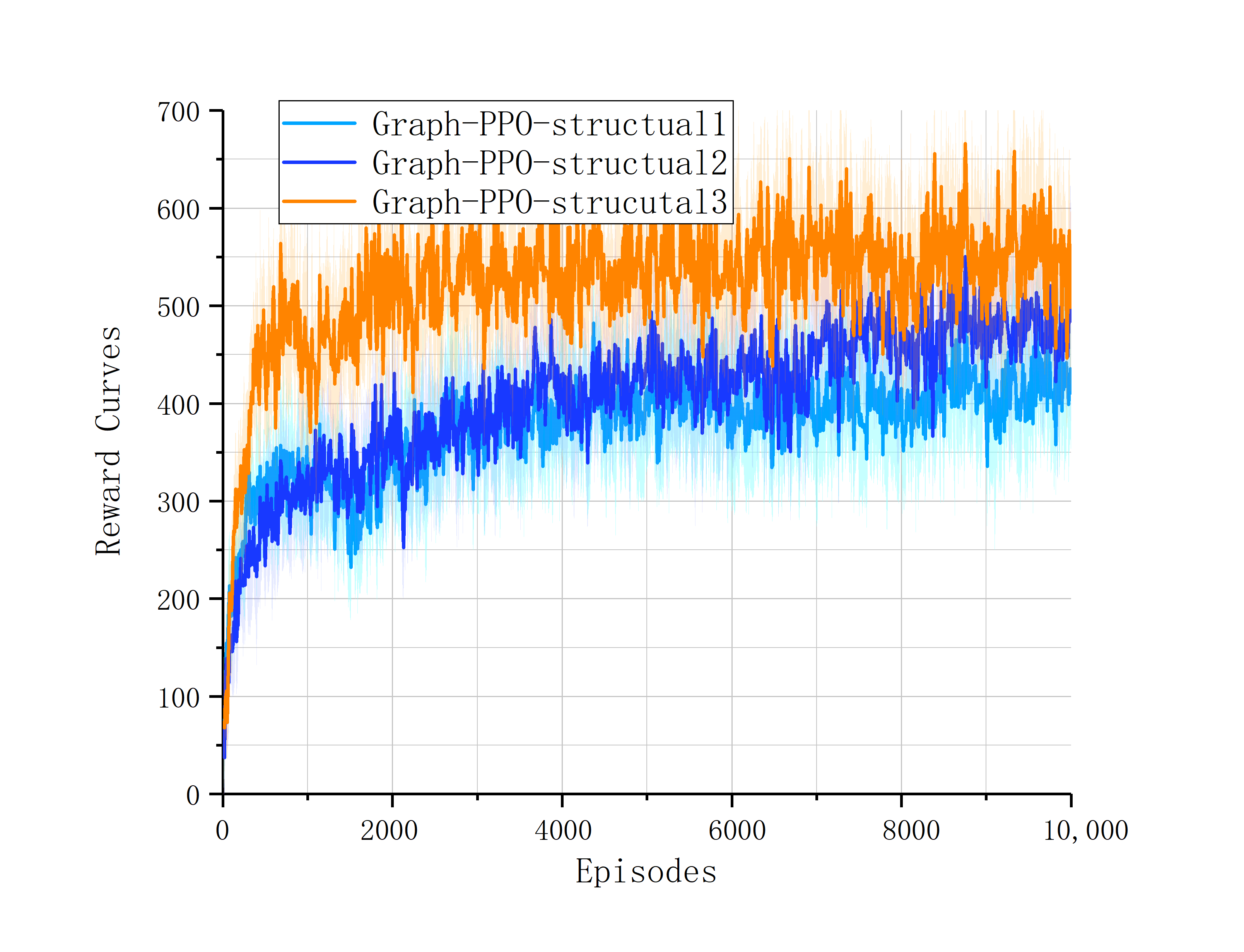}
        \caption{Reward Curves with Graph Aggregator}  
        \label{reward456_curves}
    \end{minipage}
\end{figure}

By analyzing the reward curves in Fig.~\ref{reward123_curves}, we observe that within the PPO Actor-Critic framework, a Critic that has access to the global state of the agents contributes more effectively to learning a better policy than a Critic that only utilizes individual agent states. In terms of convergence performance, the PPO variant with an individual-state Critic converges at approximately 380, whereas the PPO variant with a global-state Critic converges at around 400. Notably, the PPO architecture with a Shared Backbone Actor-Critic achieves the best performance, with the reward curve converging at approximately 450.
Fig.~\ref{reward456_curves} further presents the results after incorporating the Graph Aggregator mechanism into the PPO Actor-Critic architecture. The Shared Backbone PPO architecture still demonstrates a clear advantage. Specifically, the PPO model with an individual-state Critic converges at approximately 400, while the model with a global-state Critic converges at around 430. In contrast, the Shared Backbone Actor-Critic PPO achieves the highest performance, with the reward curve converging at approximately 500. These results indicate that, under both settings, the Shared Backbone Actor-Critic architecture consistently improves PPO learning performance.
By comparing the two figures, we further find that introducing the Graph Aggregator leads to better overall performance. The improvement is most significant under the Shared Backbone Actor-Critic architecture, while only marginal gains are observed in the other two architectures.


\begin{figure}[htbp]
    \begin{minipage}[t]{0.5\textwidth}
        \centering
        \includegraphics[width=7cm]{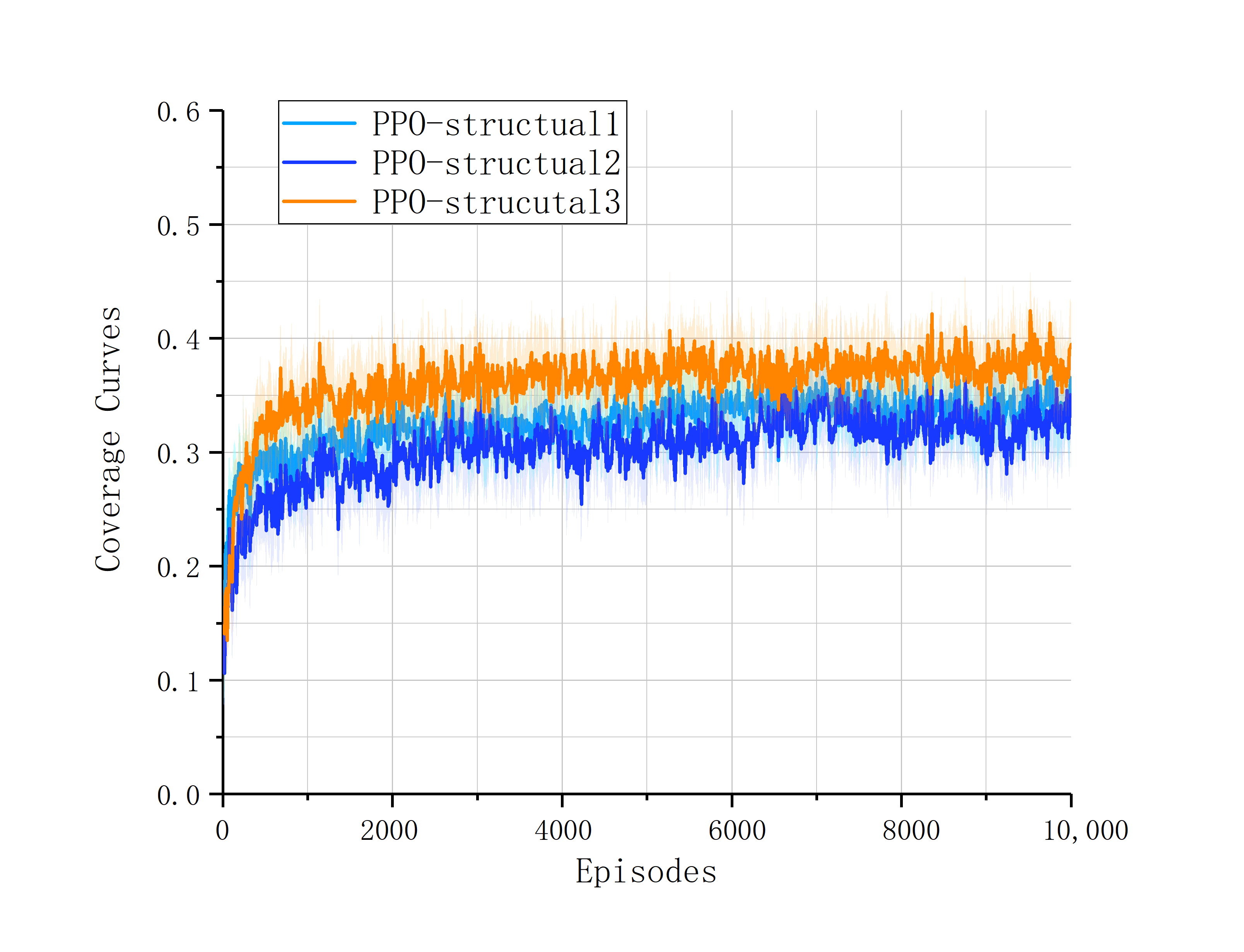}
        \caption{Coverage Curves without Graph Aggregator} 
        \label{coverage123_curves}
    \end{minipage}
    \begin{minipage}[t]{0.5\textwidth}
        \centering
        \includegraphics[width=7cm]{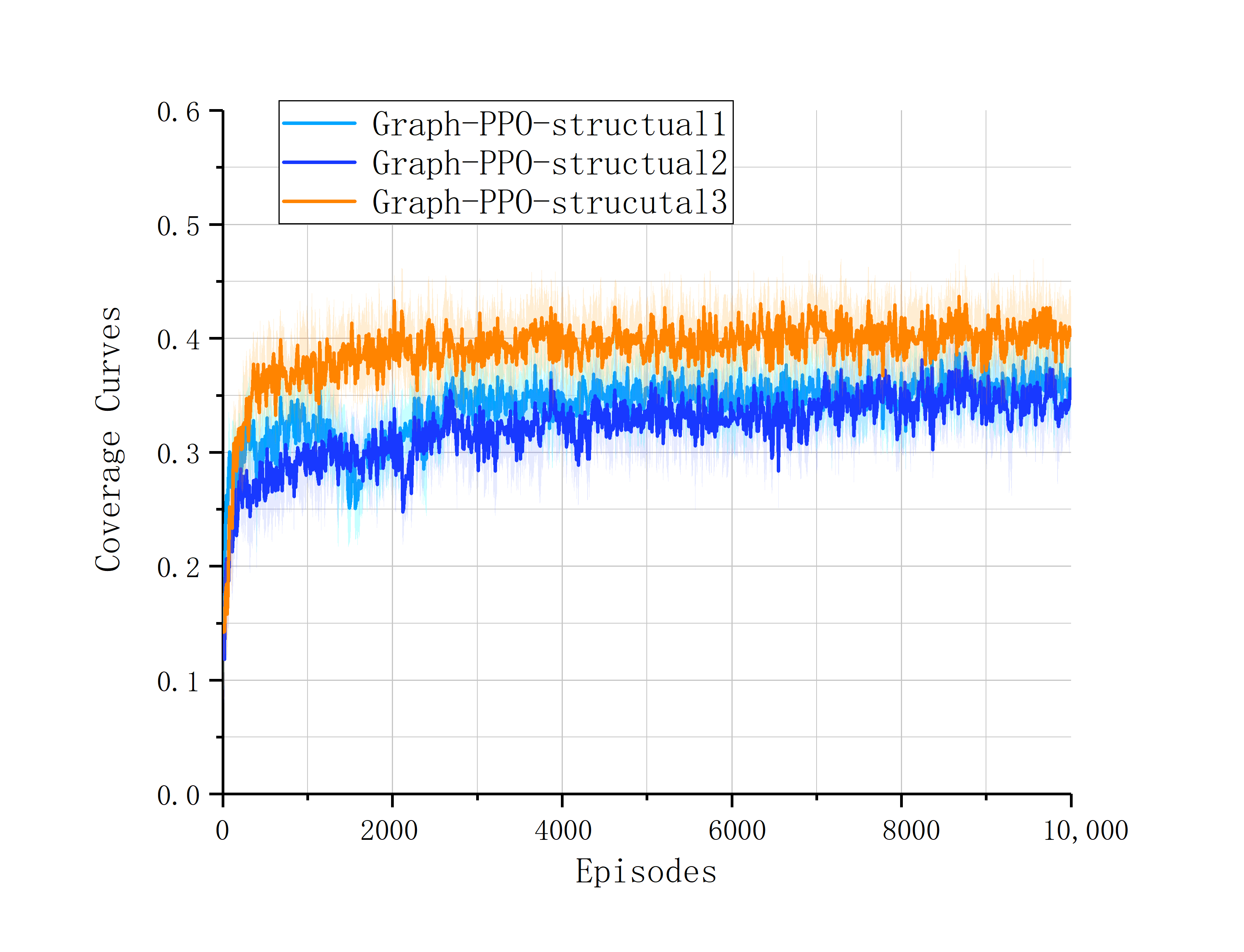}
        \caption{Coverage Curves with Graph Aggregator} 
        \label{coverage456_curves}
    \end{minipage}
\end{figure}

Fig.~\ref{coverage123_curves} and Fig.~\ref{coverage456_curves} illustrate the coverage rate curves of the UAV swarm, which are used to evaluate the degree of coverage over ground terminal nodes. The coverage index is defined such that it equals 1 when all terminal nodes are fully covered, and 0 when no nodes are covered.
From the experimental results, we observe that all PPO variants exhibit relatively weak performance at the beginning of training. However, after approximately 500 episodes, the performance of all models improves significantly. By around 1000 episodes, the learning process gradually stabilizes, and the performance becomes largely convergent. During the training process between 1000 and 2000 episodes, the PPO model with a global Critic and Graph Aggregator experiences a temporary performance drop, but it subsequently recovers and returns to a high-performance level.
Overall, among all compared methods, the PPO variant with a global Critic consistently outperforms the one with an individual Critic. Similarly, models equipped with the Graph Aggregator achieve better performance than those without graph aggregation. The Shared Backbone Actor-Critic PPO achieves the best overall coverage performance among all architectures. In terms of performance improvement, the combination of the Graph Aggregator and the Shared Backbone Actor-Critic design yields the most significant gains.


\begin{figure}[htbp]
    \begin{minipage}[t]{0.5\textwidth}
        \centering
        \includegraphics[width=7cm]{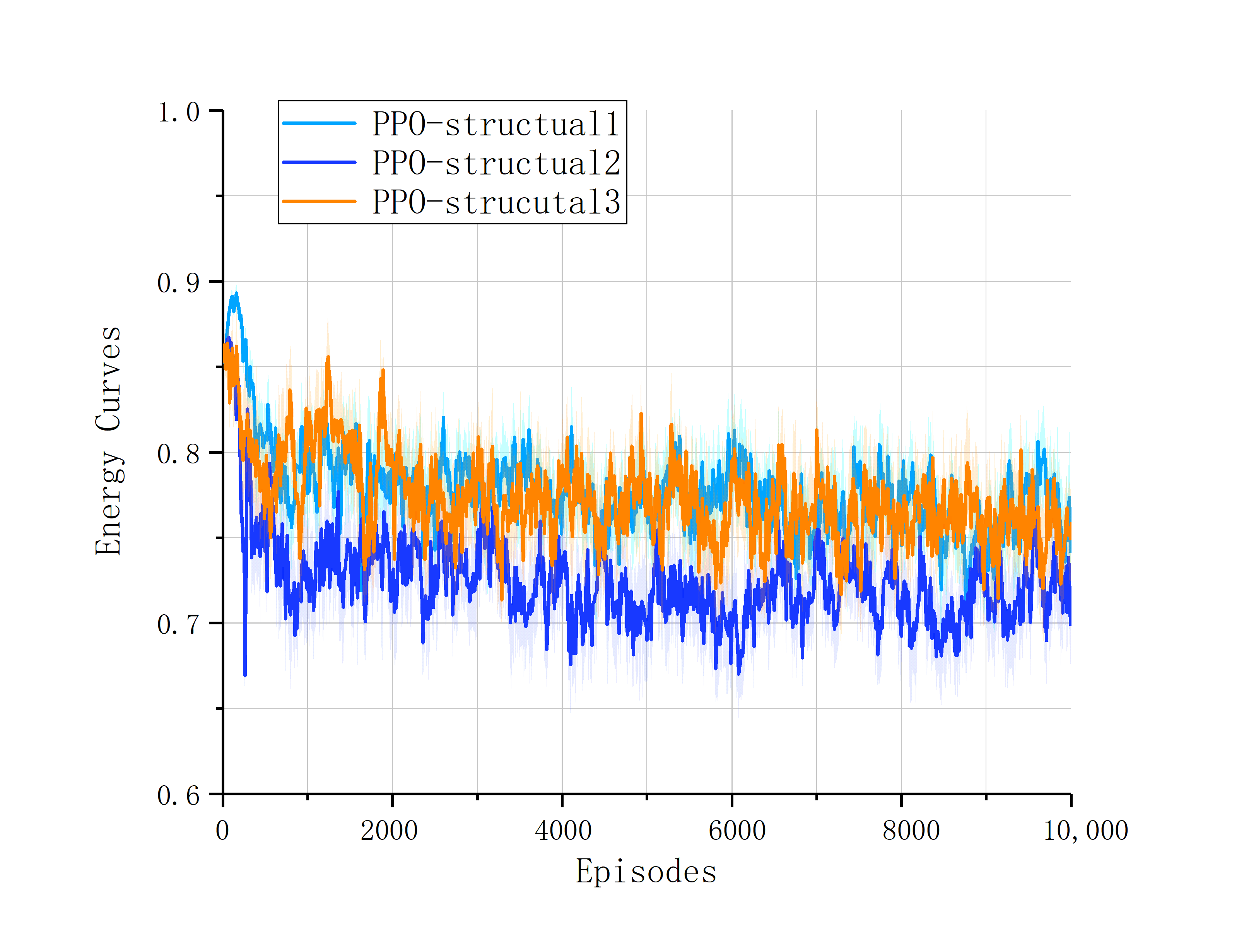}
        \caption{Energy Curves without Graph Aggregator}
        \label{energy123_curves}
    \end{minipage}
    \begin{minipage}[t]{0.5\textwidth}
        \centering
        \includegraphics[width=7cm]{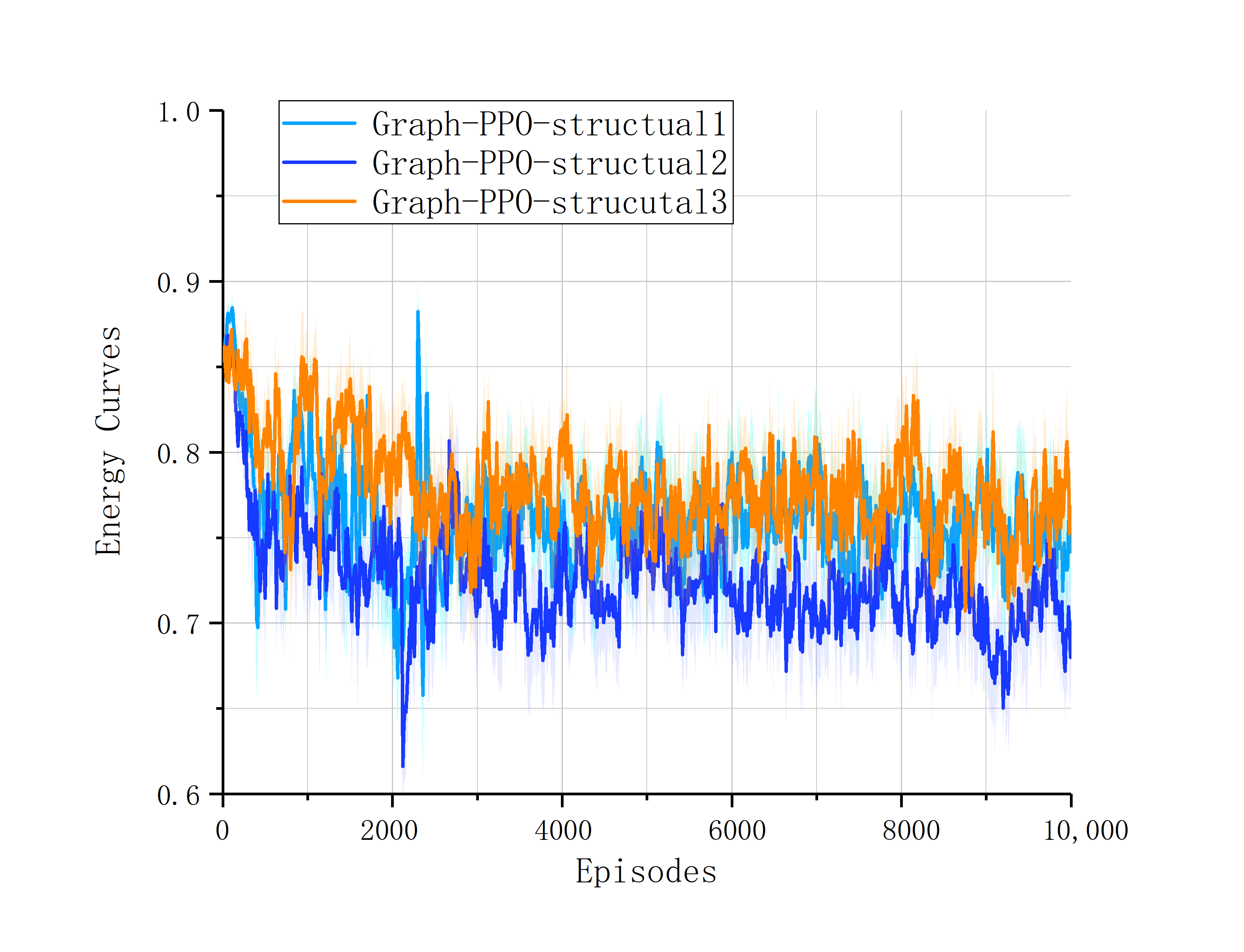}
        \caption{Energy Curves with Graph Aggregator}
        \label{energy456_curves}
    \end{minipage}
\end{figure}

\begin{figure}[htbp]
    \begin{minipage}[t]{0.5\textwidth}
        \centering
        \includegraphics[width=6cm]{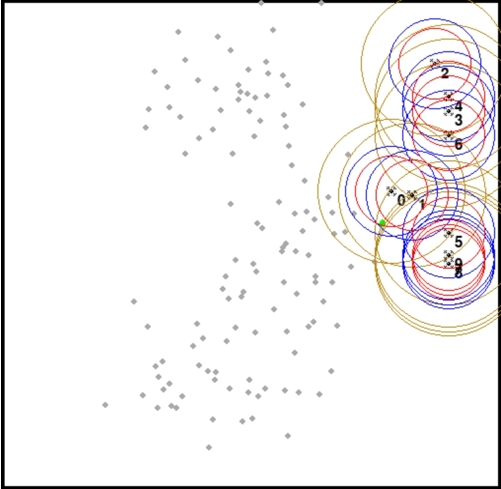}
        \caption{Coverage before Train}
        \label{coverage-before-train}
    \end{minipage}
    \begin{minipage}[t]{0.5\textwidth}
        \centering
        \includegraphics[width=6cm]{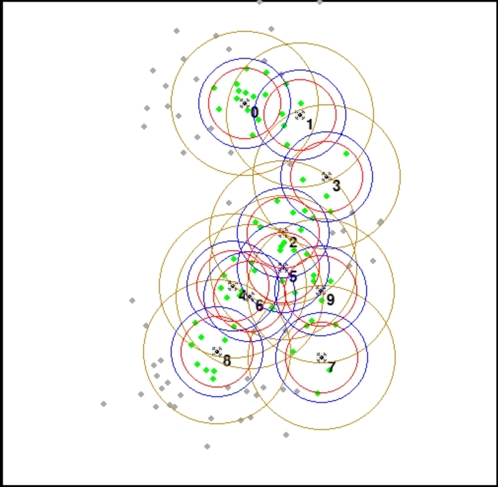}
        \caption{Coverage after Train}
        \label{coverage-after-train}
    \end{minipage}
\end{figure}

Next, we analyze the energy consumption of the UAV swarm, as shown in Fig.~\ref{energy123_curves} and Fig.~\ref{energy456_curves}. The energy index is defined as a metric for UAV flight energy consumption, where full-speed forward flight corresponds to an energy cost of 1, while near-stationary hovering to provide services for terminal nodes corresponds to an energy cost of 0.5. The reported value is computed as the average over 100 time steps within each episode.
From the results in the figures, we observe that as training progresses, the algorithms gradually learn more energy-efficient flight strategies, leading to a continuous decrease in energy consumption. After approximately 2000 episodes, the energy consumption curves become largely stable and converge.
From the comparison, we find that the Shared Backbone Actor-Critic PPO and the PPO variant with an individual-state Critic achieve comparable performance in terms of energy efficiency. In contrast, the PPO with a global-state Critic demonstrates superior performance in reducing the energy consumption of the UAV swarm.
Fig.~\ref{coverage-before-train} and Fig.~\ref{coverage-after-train} illustrate the task performance of the Shared Backbone PPO before and after training in UAV swarm network coverage tasks.


\section{Conclusion}
In this paper, we introduce a structural modification to the original PPO Actor-Critic framework, where the originally independent Actor and Critic networks share a common backbone. Compared with the standard PPO architecture with separate Actor and Critic networks, the proposed Shared Backbone PPO demonstrates improved performance.
Meanwhile, we incorporate a Graph Aggregator module into the architecture. By embedding the Graph Aggregator, the PPO agent is able to aggregate information from neighboring agents, enabling more effective policy decision-making and addressing the limitation of conventional PPO methods in handling inter-agent communication in multi-agent settings.
We apply the proposed method to a multi-UAV cooperative communication coverage scenario and compare its performance with the baseline PPO and several modified PPO variants. The experimental results show that introducing the Graph Aggregator can improve certain aspects of policy performance, while adopting a Shared Backbone Actor-Critic structure leads to a more significant enhancement in decision-making capability.

Under the same limited number of training episodes and time steps, the Actor branch in the Shared Backbone PPO can fully leverage information from neighboring agents for decision inference, while the Critic branch focuses more precisely on the local state of the current agent for targeted value estimation. As a result, the training process achieves more stable convergence and leads to improved cooperative behavior among agents.

In the future, we also plan to validate the proposed method on real UAV platforms and conduct a series of decision-making experiments in physical environments, enabling the learned policies to better adapt to real-world scenarios and create practical value for real deployment tasks.


\newpage


\begin{thebibliography}{99}

\bibitem{shao2024deepseekmath}
Shao, Z., Wang, P., Zhu, Q., Xu, R., Song, J., Bi, X., Zhang, H., Zhang, M., Li, Y. K., Wu, Y., et al. (2024). 
Deepseekmath: Pushing the limits of mathematical reasoning in open language models. 
\textit{arXiv preprint arXiv:2402.03300}.

\bibitem{alam2022survey}
Alam, M. M., \& Moh, S. (2022). 
Survey on Q-learning-based position-aware routing protocols in flying ad hoc networks. 
\textit{Electronics, 11}(7), 1099.

\bibitem{wu2022deep}
Wu, S., Pu, Z., Qiu, T., Yi, J., \& Zhang, T. (2022). 
Deep-reinforcement-learning-based multitarget coverage with connectivity guaranteed. 
\textit{IEEE Transactions on Industrial Informatics, 19}(1), 121--132.

\bibitem{rezwan2021survey}
Rezwan, S., \& Choi, W. (2021). 
A survey on applications of reinforcement learning in flying ad-hoc networks. 
\textit{Electronics, 10}(4), 449.

\bibitem{pasandideh2022review}
Pasandideh, F., da Costa, J. P. J., Kunst, R., Islam, N., Hardjawana, W., \& Pignaton de Freitas, E. (2022). 
A review of flying ad hoc networks: Key characteristics, applications, and wireless technologies. 
\textit{Remote Sensing, 14}(18), 4459.

\bibitem{jiang2023graph}
Jiang, Z., Chen, Y., Wang, K., Yang, B., \& Song, G. (2023). 
A graph-based PPO approach in multi-UAV navigation for communication coverage. 
\textit{International Journal of Computers Communications \& Control, 18}(6).

\bibitem{schulman2017proximal}
Schulman, J., Wolski, F., Dhariwal, P., Radford, A., \& Klimov, O. (2017). 
Proximal policy optimization algorithms. 
\textit{arXiv preprint arXiv:1707.06347}.

\bibitem{haarnoja2018soft}
Haarnoja, T., Zhou, A., Abbeel, P., \& Levine, S. (2018). 
Soft actor-critic: Off-policy maximum entropy deep reinforcement learning with a stochastic actor. 
In \textit{International Conference on Machine Learning}, pp. 1861--1870. PMLR.

\bibitem{watkins1992q}
Watkins, C. J. C. H., \& Dayan, P. (1992). 
Q-learning. 
\textit{Machine Learning, 8}, 279--292.

\bibitem{mnih2013playing}
Mnih, V. (2013). 
Playing Atari with deep reinforcement learning. 
\textit{arXiv preprint arXiv:1312.5602}.

\bibitem{jiang2024fault}
Jiang, Z., Song, T., Yang, B., \& Song, G. (2024). 
Fault-tolerant control for multi-UAV exploration system via reinforcement learning algorithm. 
\textit{Aerospace, 11}(5), 372.

\bibitem{guo2025deepseek}
Guo, D., Yang, D., Zhang, H., Song, J., Zhang, R., Xu, R., Zhu, Q., Ma, S., Wang, P., Bi, X., et al. (2025). 
Deepseek-r1: Incentivizing reasoning capability in LLMs via reinforcement learning. 
\textit{arXiv preprint arXiv:2501.12948}.

\bibitem{jiang2023cooperative}
Jiang, Z., Chen, Y., Song, G., Yang, B., \& Jiang, X. (2023). 
Cooperative planning of multi-UAV logistics delivery by multi-graph reinforcement learning. 
In \textit{International Conference on Computer Application and Information Security (ICCAIS)}, pp. 129--137. SPIE.

\bibitem{han2025differentiated}
Han, Y., Zhang, L., \& Meng, D. (2025). 
A differentiated reward method for reinforcement learning based multi-vehicle cooperative decision-making algorithms. 
\textit{arXiv preprint arXiv:2502.00352}.

\bibitem{ray2019benchmarking}
Ray, A., Achiam, J., \& Amodei, D. (2019). 
Benchmarking safe exploration in deep reinforcement learning. 
Available at: https://cdn.openai.com/safexp-short.pdf

\bibitem{zhang2024user}
Zhang, Z., Zhang, Q., Wu, X., Shi, X., Liao, G., Wang, Y., Wang, X., \& Zhao, D. (2024). 
User response modeling in reinforcement learning for ads allocation. 
In \textit{Companion Proceedings of the ACM Web Conference}, pp. 131--140.

\bibitem{bai2023towards}
Bai, Y., Zhao, H., Zhang, X., Chang, Z., J\"antti, R., \& Yang, K. (2023). 
Towards autonomous multi-UAV wireless network: A survey of reinforcement learning-based approaches. 
\textit{IEEE Communications Surveys \& Tutorials}.

\bibitem{hosseinzadeh2023novel}
Hosseinzadeh, M., Ali, S., Ionescu-Feleaga, L., Ionescu, B. S., Yousefpoor, M. S., Yousefpoor, E., Ahmed, O. H., Rahmani, A. M., \& Mehmood, A. (2023). 
A novel Q-learning-based routing scheme using an intelligent filtering algorithm for flying ad hoc networks (FANETs). 
\textit{Journal of King Saud University - Computer and Information Sciences, 35}(10), 101817.

\bibitem{arafat2021q}
Arafat, M. Y., \& Moh, S. (2021). 
A Q-learning-based topology-aware routing protocol for flying ad hoc networks. 
\textit{IEEE Internet of Things Journal, 9}(3), 1985--2000.

\bibitem{sutton2018reinforcement}
Sutton, R. S. (2018). 
\textit{Reinforcement learning: An introduction}. A Bradford Book.

\bibitem{cao2021novel}
Cao, L., Yue, Y., Cai, Y., \& Zhang, Y. (2021). 
A novel coverage optimization strategy for heterogeneous wireless sensor networks based on connectivity and reliability. 
\textit{IEEE Access, 9}, 18424--18442.

\bibitem{trotta2020gps}
Trotta, A., Montecchiari, L., Di Felice, M., \& Bononi, L. (2020). 
A GPS-free flocking model for aerial mesh deployments in disaster-recovery scenarios. 
\textit{IEEE Access, 8}, 91558--91573.

\end{thebibliography}
\end{document}